\documentclass{tlp}
\usepackage{amsmath}
\usepackage{amssymb}
\usepackage{epsfig}
\usepackage{graphics}
\usepackage{graphicx}
\usepackage{moreverb}
\usepackage{color}

\def\dom{\mbox{\rm dom}}
\def\true{\mbox{\it true}}
\def\false{\mbox{\it false}}
\def\path{\mbox{\rm path}}
\def\sol{\mbox{\rm sol}}

\begin{document}

\title[Inference with Constrained Hidden Markov Models]
{Inference with Constrained Hidden Markov Models in PRISM}

  \author[H. Christiansen et al.]
         {Henning Christiansen, Christian Theil Have, Ole Torp Lassen and Matthieu Petit\\
    Research group PLIS: Programming, Logic and Intelligent Systems\\
Department of Communication, Business and Information Technologies\\
Roskilde University, P.O.Box 260, DK-4000 Roskilde, Denmark\\
         \email{\{henning, cth, otl, petit\}@ruc.dk}}

\newtheorem{definition}{Definition}[section]
\newtheorem{proposition}{Proposition}[section]

\maketitle

\begin{abstract}
A Hidden Markov Model (HMM) is a common statistical model which is widely used for analysis of biological sequence data and other sequential phenomena. In the present paper 
we show how HMMs can be extended with side-constraints and present constraint solving techniques for efficient inference.
Defining HMMs with side-constraints 
in Constraint Logic Programming have advantages in terms of more compact expression and pruning opportunities during inference. 
We present a PRISM-based framework for extending HMMs with side-constraints and show how well-known constraints such as \texttt{cardinality} and \texttt{all\_different}
are integrated. We experimentally validate our approach on the
biologically motivated problem of global pairwise alignment. \\

\begin{keywords}
Hidden Markov Model with side-constraints, Inference, PRogramming In Statistical Modeling
\end{keywords}

\textbf{Note:} This article has been published in Theory and Practice of Logic Programming,
10(4-6), 449–464, \copyright Cambridge University Press.
\end{abstract}

\section{Introduction}
Hidden Markov Models (HMMs) are one of the most popular models for analysis
of sequential processes taking place in a random way, where ``randomness'' may 
also be an abstraction covering the fact that a detailed analytical model for 
the internal matters is unavailable. Such a sequential process can be
observed from outside by its
emission sequence (letters, sounds, measures of features, all kinds of signals) produced
over time, and an HMM postulates a hypothesis about the internal machinery in terms of
a finite state automaton equipped with probabilities for the different state transitions
and single  emissions.
A common inference for a given observed sequence
means to compute the ``best'' state transitions that the HMM may go through
to produce the sequence, and thus this represents a best hypothesis for the
internal structure or ``content'' of the sequence.
HMMs are widely used in speech recognition and biological sequence analysis~\cite{Rab89,DEK98}.

The efficiency of computations on HMMs heavily depends on the Markov property.
Decisions made during a process run depends only on a limited past. 
Dynamic programming algorithms, such as Viterbi and Forward-Backward, are then used to 
perform efficient inference. However, many problems would require more
complex dependencies among elements of the process.
For example, it may be interesting
to constrain an HMM to visit only different states or limit the number of visits
to a given state.
It is possible to model  the \texttt{all\_different} constraint for the states visited by
extending the underlying finite state automaton,
but for the price of a factorial number of 
new states and with an obvious impact on inference.
As an alternative to modifying the HMM structure, we instead extend the HMM with side-constraints~\cite{SK08,RY05}. 
However, classical algorithms, such as Viterbi, must be modified to take 
care about these side-constraints \cite{CRR08,CHL09}.

In this paper, we
extend HMMs with side-constraints, leading to what we call Constrained HMMs (CHMMs).
Side-constraints are external constraints declared in addition to those defined by
the structure of an HMM. The concept of CHMMs was introduced by Sato et al. in~\cite{SK08},
although earlier and unrelated systems have used the same or similar names
(discussed in section \ref{sec:related_work}).
The contribution of this paper is to define CHMMs as  constraint logic programs extended with probabilistic
choices and to show how to employ this setting for more efficient Viterbi computation, i.e., computation of the most
probable explanation of an observation.
Moreover, defining HMMs with side-constraints in Constraint Logic Programming have advantages in terms of more compact 
expression and pruning opportunities during inference.
We show how to implement CHMMs in PRISM \cite{SK97} and 
how to integrate well-known constraints, such as \texttt{cardinality} and \texttt{all\_different}, into this framework.
We validate our approach experimentally on the biologically motivated problem of global pairwise alignment.

The paper is organized as follows:
section \ref{sec:background} describes background on HMMs.  
In section \ref{sec:constraint_model}, we formally introduce the constraint model associated with a CHMM. Section \ref{sec:implementation_CHMM}
describes our PRISM-based framework to define CHMMs. Section \ref{sec:experiment_validation} presents an experimental validation. 
Finally, sections \ref{sec:related_work} and \ref{sec:discussion} present related work and conclusions. 

\vspace{-0.2cm}

\section{Background}\label{sec:background}

Here we define Hidden Markov Models (HMM)s and illustrate their application
to the problem of pairwise global alignment.

\subsection{Hidden Markov Models}\label{sec:hidden_markov_models}

For simplicity of the technical definitions, we limit 
ourselves to a discrete Hidden Markov Model  with a distinguished
initial state.

\begin{definition}\label{def:hmm}
A \emph{Hidden Markov Model} (HMM) is 
a 4-tuple $\langle S, A, T, E\rangle$, where
\begin{itemize}
\item $S=\{s_0,s_1,\ldots,s_m\}$ is a set of \emph{states} which includes an \emph{initial} state
referred to as $s_0$;
\item $A=\{e_1,e_2,\ldots,e_k\}$ is a finite set of emission symbols;
\item $T=\left\{(p(s_0; s_1),\ldots,p(s_0; s_m)),\ldots,(p(s_m; s_1),\ldots,p(s_m; s_m))\right\}$ is a set of \emph{transition probability distributions}   
representing probabilities to transit from one state to another;
\item $E=\left\{(p(s_1 ; e_1),\ldots,p(s_1 ; e_k)),\ldots,(p(s_m ; e_1),\ldots,p(s_m ; e_k))\right\}$ is a set of \emph{emission probability distributions}
representing probabilities to emit each symbol from each state. 
\end{itemize}
We define a \emph{run} of an HMM as a pair consisting of a sequence of states 
\small $s^{(0)} s^{(1)} \ldots s^{(n)}$\normalsize, called a \emph{path} and a corresponding sequence of
emissions \small $e^{(1)} \ldots  e^{(n)}$\normalsize,  called an \emph{observation},
such that
\begin{itemize}
\item $s^{(0)}=s_0$;
\item $\forall i,0\leq i \leq n-1,p(s^{(i)};s^{(i+1)})> 0$ (probability to transit from $s^{(i)}$ to $s^{(i+1)}$);
\item $\forall i,0<i\leq n, p(s^{(i)};e^{(i)}) > 0$ (probability to emit $e^{(i)}$ from $s^{(i)}$).\end{itemize}
The \emph{probability} of such a run is defined as
$\prod_{i=1..n}p(s^{(i-1)};s^{(i)})\cdot p(s^{(i)};e^{(i)})$.

\end{definition}
%
%
%
%


\subsection{An example HMM: pairwise global alignment}\label{sec:example}

As an example of an HMM that we later extend with constraints,
we consider the problem of aligning two sequences.
Sequence alignment is among the most common
tasks in computational biology, where it is used to align
sequences assumed to have diverged from a common ancestor.
Notice that we here use a so-called pair HMM~\cite{DEK98} which emits two
sequences at the same time, and which is a straightforward
extension of the definition above.

In the global alignment problem, two sequences $x$ and $y$ must be
aligned optimally, based on a scoring scheme for comparison 
of different alignments. In probabilistic modeling,
a probability is associated with each pair of symbols emitted from a state and 
similarly a probability for introducing
gaps, $\delta$, and extending gaps, $\epsilon$, in the alignment of
the sequences is defined. The probability of an alignment is then the
product of probabilistic transitions performed to recognize the alignment.
In biology, these probabilities are defined to reflect observed statistics about sequence
mutations and conservation.

\begin{figure}[htb]
\vspace{-0.3cm}
\centerline{
\includegraphics[scale=0.35]{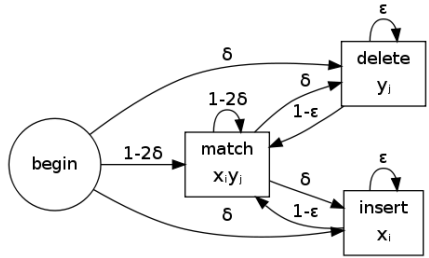}
}
\caption{A pair HMM for pairwise global alignment of sequences.
States, represented by squares for emitting states and circles for silent states, are connected by arrows representing
transitions labeled with probabilities.
}\label{fig:pairhmm}
\end{figure}

Fig. \ref{fig:pairhmm} shows an HMM capable of generating a pair of aligned sequences. When
given two sequences to align, then a path from the initial state, \texttt{begin}, such that the 
model emits the two sequences, corresponds to an alignment. The initial state, \texttt{begin}, 
does not emit symbols. The \texttt{match} state emits a pair of symbols ($x_i,y_j$), one for each
sequence corresponding to alignment of the symbol at position $i$ in
sequence $x$ and the symbol at position $j$ in sequence $y$. Emitted symbols
can be identical or different. A difference represents a potential mutation between
the two sequences. The \texttt{insert} state emits only the next symbol of sequence $x$,
effectively aligning position $x_{i}$ to a gap in $y$.
Oppositely, the \texttt{delete} state aligns a symbol $y_{j}$ to a gap in sequence $x$.

The following example shows an alignment of two
short protein sequences, where the third line indicates the state sequence
of this alignment abbreviated with the first letter of the state name:
\begin{verbatim}
  Sequence x:   H G K K G A     A Q V
  Sequence y:         K G P K K A Q A
  alignment : b i i i m m m d d m m m 
\end{verbatim}
In this context, a common task is to find
the optimal alignment. This means to find a state sequence  
that can recognize the two sequences and has maximal probability.
Another is to calculate the probability to observe an
emission sequence.
A third type of inference is parameter learning, where we are given a set
of alignments and estimate the ``best'' parameters for the model, where best
usually means that they maximize likelihood of the alignments.

\vspace{-0.2cm}
\section{A constraint model for HMM with side-constraints}\label{sec:constraint_model}

In this section, we give a formal definition of CHMMs and propose a constraint model for CHMM runs.
Then, the computation of the most probable path is adapted for CHMMs.

 \subsection{Constrained Hidden Markov Model}
A CHMM extends an HMM with constraints that limit the
set of valid runs and leave fewer paths to consider for any given sequence.

\begin{definition}
A \emph{constrained HMM (CHMM)} is defined by a 5-tuple
\small $\left\langle S,A,T,E,C\right\rangle$ \normalsize where
\small $\left\langle S, A,T,E\right\rangle$ \normalsize is an HMM and
$C$ is a set of constraints, each of which is a mapping from 
HMM runs into $\{true, false\}$.

A \emph{run} of a CHMM, $\langle path,observation\rangle$ is a run of the corresponding
HMM for which $C(path,observation)$ is true.
\end{definition}
Notice that we define constraints in a highly abstract way, independently of
any specific constraint language. In the following, constraints over 
finite domains \cite{VSD95} are used, although other constraint languages such as $CLP(Q)$
and $CLP(R)$ could have been used as well.

\subsection{Runs of a CHMM as a constraint program}

In this section, we propose to model runs of CHMM by a constraint program over finite domains. In this context, a run of CHMM is a solution of
the constraint program.

Let $\left\langle S, A,T,E, C\right\rangle$ be a CHMM
and $n$ the sequence length.
A constraint program for runs is given by the following predicate.
\begin{displaymath}
run([s^{(0)},S_1,\ldots,S_{n}],[E_1,\ldots,E_n])
\end{displaymath}
where each variable $S_i$ and $E_i$ represents the state and the emission at the step $i$.
The domains of $S_i$ and $E_i$,
are given as $\dom(S_i)=S\setminus \{s_0\}$
and $\dom(E_i)=E$.
The $run$ predicate is specified as follows.

\medskip
$run([s^{(0)},S_1,\ldots,S_{n}],[E_1,\ldots,E_n])$ is $\true$ iff 
\begin{multline}\label{eq:constraint_model}
\exists s^{(1)}\in \dom(S_1),\ldots,\exists s^{(n)}\in\dom(S_n) \text{ and } \\
\exists e^{(1)}\in\dom(E_1),\ldots,\exists e^{(n)}\in\dom(E_n), \\
C(s^{(0)} s^{(1)}\ldots s^{(n)},e^{(1)}\ldots e^{(n)}) \text{ is true}, s^{(0)}=s_0 \text{ and}\\
p(s^{(0)};s^{(1)})\cdot p(s^{(1)};e^{(1)})\ldots p(s^{(n-1)};s^{(n)})\cdot p(s^{(n)};e^{(n)})>0. 
\end{multline}
Formula~(\ref{eq:constraint_model}) states that $s^{(0)} s^{(1)}\ldots s^{(n)}$ and $e^{(1)}\ldots e^{(n)}$ is a run of the HMM
that satisfies $C$. 
By the definition of \emph{run/2}, (local) relationships between $S_i$ and $S_{i+1}$ and  $S_i$ and $E_{i}$ can be established, 
since the probability of a run must be positive. Indeed, valuation of $S_i$ to $s^{(i)}$ and $S_{i+1}$ to $s^{(i+1)}$
can be part of a solution of the constraint program whenever $p(s^{(i)};s^{(i+1)})>0$. 
These relationships between variables of $run/2$ are modeled by the following constraints, \begin{displaymath}
trans(S_{i-1},S_{i}) \textrm{ and } emit(S_i,E_{i}), \text{for all $i$, } 1\leq i \leq n
\end{displaymath}
where $S_i$, $S_{i+1}$ and $E_{i}$ are the variables of \emph{run\slash 2}.
These constraints are defined as follows.
\begin{itemize}
\item
$trans(S_i, S_{i+1})$ is $\true$ iff 
\small $\exists s^{(i)}\in\dom(S_i)\text{ and } s^{(i+1)}\in\dom(S_{i+1}) \text{ such that } \\ p(s^{(i)};s^{(i+1)})>0$; \normalsize 
\item
$emit(S_i, E_i)$ is $\true$ iff 
\small $\exists s^{(i)}\in\dom(S_i)\text{ and }e^{(i)}\in\dom(E_i) \text{ such that } p(s^{(i)};e^{(i)})>0$. \normalsize  
\end{itemize}
Section \ref{sec:implementation_CHMM} below shows an implementation of
this framework such that
a solution of the constraint program corresponds to a valid derivation of a PRISM program.

\subsection{Example: constrained pairwise global alignment}
We consider the HMM presented in section \ref{sec:example} and
extend it into a CHMM by the following
set of constraints,
\begin{multline*}
C= \left\{\texttt{cardinality\_atmost}(N_d,[S_1,\ldots,S_{n}],\text{delete}),\right.\\
\left. \texttt{cardinality\_atmost}(N_i,[S_1,\ldots,S_{n}],\text{insert})\right\}.
\end{multline*}
A constraint \texttt{cardinality\_atmost}$(N,L,X)$ is satisfied whenever $L$
is a list of elements, out of which at most $N$ are equal to $X$.
In a biological context, it is reasonable to consider only alignments
with a limited number of insertions and deletions given the assumption that the two sequences are related.

As described above, we can consider this CHMM as a constraint program
\begin{displaymath}
run([s^{(0)},S_1,\ldots,S_{n}],[E_1,\ldots,E_n])
\end{displaymath}
where $\dom(S_i)\in\{\text{match},\text{delete},\text{insert}\}$,
$\dom(E_i)\in\{A,C,D,\ldots,W,Y\}$\footnote{This set of letters
refers to the 21 different amino acids from which proteins are composed.}
and the constraints $C$ are as described above.

\subsection{Computation of the most probable path for a CHMM}\label{sec:Viterbi_ctr}

The Viterbi algorithm \cite{Vit67} is a dynamic programming algorithm for
finding a most probable path corresponding to a given observation. The algorithm 
keeps track of, for each prefix of an observed emission sequence,
the most probable (partial) path leading to each possible state, and extends
those step by step into longer paths, eventually covering the entire emission sequence.
Here, we adapt this algorithm for CHMMs.

Consider a given observation $e^{(1)}\ldots e^{(n)}$,
a CHMM $\langle S,A,T,E,C\rangle$,  and
its constraint program
\begin{displaymath}
run([s^{(0)},S_1,\ldots,S_{n}],[e^{(1)},\ldots,e^{(n)}]).
\end{displaymath}
The most probable path is computed by finding the valuation $s^{(1)},\ldots,s^{(n)}$ 
that maximizes the objective function: the probability of a run.

Computation of the most probable path for CHMM is expressed as a rewriting system on a set of 5-tuples $\Sigma$.
Each such 5-tuple is of form $\langle s,i,p,\pi,\sigma\rangle$
where $\pi$ is a partial path ending in state $s$
and representing a path for the emission sequence prefix
$e^{(1)}\cdots e^{(i)}$; $p$ is the computed probability for the emissions and transitions applied
in the construction of $\pi$, and $\sigma$ is the current constraint store seen
as a conjunction of constraints.
Any ground and satisfied constraint will be removed from the constraint store,
and $true$ refers to the empty conjunction.
The set of solutions of a constraint store $\sigma$ is denoted by $\sol(\sigma)$.

The two rewriting rules in Fig.~\ref{fig:Viterbi_CHMM} describe an iteration step
of the computation of the most probable path.\footnote{When any reference to
 constraints and the constraint store are removed from Fig.~\ref{fig:Viterbi_CHMM},
 we have a compact representation of one iteration step of the Viterbi algorithm for HMMs.}
The computation starts from an initial set of 5-tuples
\begin{multline}\label{eq:init_state}
\{ \langle s^{(0)},0,1,\epsilon,C\wedge trans(s^{(0)},S_1)\wedge  \\
\bigwedge_{1 \leq i\leq n-1} trans(S_i,S_{i+1}) \wedge \bigwedge_{1 \leq i\leq n} emit(S_i,e_i)\rangle\}.
\end{multline}
\begin{figure}
\begin{center}
\begin{tabular}{ll}
\hline &\\
   $\quad trans\_ctr:$ &
       $\Sigma:= \Sigma\cup\{\langle s',i\!+\!1,p \cdot p(s;s')\cdot p(s';e^{(i+1)}), \pi\,s',\sigma\wedge S_{i+1}=s'\rangle\}$\\
   & whenever $\langle s,i,p, \pi,\sigma\rangle\in \Sigma$, $p(s;s'),p(s';e^{(i+1)})>0\quad$ \\
   & check\_constraints($\sigma\wedge S_{i+1}=s'$) and $prune\_ctr$ does not apply.
   \\ & \\
    $\quad prune\_ctr :$ & $\Sigma:= \Sigma\setminus \{ \langle s,i\!+\!1,p', \pi',\sigma'\rangle \}$\\
    & whenever there is another $ \langle s,i\!+\!1,p, \pi,\sigma\rangle \in \Sigma$ with \\
   & $p\ge p'$ and $\sol(\sigma')\subseteq \sol(\sigma)$.
   \\ & \\
\hline
\end{tabular}
\end{center}
\caption{Rewriting rules for the computation of most probable paths for CHMM}\label{fig:Viterbi_CHMM}

\end{figure}
The $trans\_ctr$ rule expands an existing partial path one step in  
directions that preserve the satisfaction of  the constraint store;
this satisfiability check
is denoted check\_constraints (and depends thus on the particular $C$).
The $prune\_ctr$ rule removes partial solutions that are not optimal
for the current observation prefix \emph{and} 
shares the same set of complete solutions with the better partial solution.
The second condition is necessary in case no partial path
contained in sol$(\sigma)$ can be extended into a full path without
violating the constraints.
We take the following correctness property for granted.
\begin{proposition}
Assume a CHMM $H$ with the notation as above and an observation
\small $Obs=e^{(1)}\cdots e^{(n)}$. \normalsize 
When the Viterbi algorithm in Fig. \ref{fig:Viterbi_CHMM} is executed from an initial set of 5-tuples
given the formula (\ref{eq:init_state}), it terminates with a set of 5-tuples $\Sigma_{final}$.
It holds that
\begin{itemize}
\item For any $ \langle s,n,p,\pi,true\rangle\in\Sigma_{final}$, $\pi$ is a most probable path for
$Obs$ ending in $s$ and with probability $p$.
\item Whenever there exists a path for $Obs$ ending in $s$, $\Sigma_{final}$
includes a 5-tuple of the form  $ \langle s,n,p,\pi,true\rangle$.
\end{itemize}
\end{proposition}
Notice that all the variables of the constraint program are valuated 
when a final state is reached,
and thus any final constraint store is equivalent to $true$
(as $trans\_ctr$ prevents any inconsistent store to arise).

The classical Viterbi algorithm is guaranteed to run in time linear
to the length of the given sequence, whereas our algorithm may in the worst
case run in exponential time; this may occur if $prune\_ctr$ cannot be applied at all.
In other words, a representation of the constraint store that allows an efficient
comparison as in ``$\sol(\sigma')\subseteq \sol(\sigma)$'' is essential
for the practicability of our algorithm.
On the other hand,
for those problems that can be formulated as a CHMM with
effective and efficient definitions of check\_constraints and
the comparison test,
the $\Sigma$ states may stay of a reasonable size.
Notice that our algorithm is still correct if we use approximations
of these tests,
more specifically, check\_constraints may occasionally return $true$ when the correct answer is
$\false$ and the opposite for the comparison.

\vspace{-0.2cm}
\section{Implementation of CHMMs in PRISM}\label{sec:implementation_CHMM}

After briefly introducing PRISM, we propose a methodology to define CHMMs in this framework. 

\subsection{A brief introduction to the PRISM system}\label{sec:prism}
PRISM~\cite{SK08} is a powerful system
for working with probabilistic-logic models,
based on an extension to Prolog with discrete random variables, called
multi-valued switches. We illustrate this with a simple example HMM with
two states \texttt{s0} and \texttt{s1}. A switch declaration,
\begin{verbatim}
values(x,O).  
\end{verbatim}
associates the named random variable x with a
set of outcomes \verb+O+. Whenever the goal
\texttt{msw(x,X)} is called from the program, then a probabilistic
choice will be made unifying \texttt{X} with an element of \verb+O+.
Switches can also be defined in a parametric form, 
\begin{verbatim}
values(emit(_),[a,b]).  % symbol emission
values(trans(_),[s0,s1]). % state transition
\end{verbatim}
where each declaration defines a family of switches, one for each
possible instance of \texttt{emit(\_)} and \texttt{trans(\_)} and each instance
is given a distinct probability distribution. This parametrization can
serve to model dependencies: in our HMM example we define the parameters 
to be the states \texttt{s0} and \texttt{s1}
(plus \texttt{init} for \texttt{trans(\_)}), thus defining 
emissions and transitions for each state with the Markov property. 
Finally, we define a logic program to implement the probabilistic
model,
\begin{verbatim}
hmm(L):- run_length(T), hmm(T,init,L).
hmm(0,_,[]).
hmm(T,State,[Emit|EmitRest]) :-
   T > 0,
   msw(trans(State),NextState),
   msw(emit(NextState),Emit),
   T1 is T-1,
   hmm(T1,NextState,EmitRest).
run_length(10).
\end{verbatim}
Here, a derivation of the goal \texttt{hmm} corresponds to what we define as a \emph{run} in
section~\ref{sec:hidden_markov_models}. 
As shown by~\cite{Sat95}, Prolog's traditional Herbrand model semantics generalizes
immediately to a probabilistic semantics when
probabilities are given for each random variable (provided that a few restrictions
are respected on how \texttt{msw} is used in the program).
Thus a PRISM program defines a probabilistic model that provides
a probability distribution for all goals that can be formulated
in the program's logical language. 
PRISM assigns each possible derivation of a goal a
probability defined as the product of the probabilities of the
selected switch outcomes of switches used in the derivation.
Under normal conditions, 
it will be the case that the sum of probabilities of all possible
derivations of such a goal is unity, but these conditions can be violated in a
constrained model. 
If a program attempts to unify the stochastically selected outcome of
a switch with some other value distinct from that outcome, this
unification will fail resulting in a failed derivation.

PRISM includes built-in  mechanisms for efficient probabilistic
inference based on tabling. During inference, once a probabilistic goal
has been solved, its answers are put in a global table.
Later calls to the same goal will simply lookup the answer
in the table in constant time. PRISM utilizes this to provide 
an efficient generalized Viterbi algorithm that may be used for
the computation of the most likely \emph{successful} derivation for
a large number of probabilistic models including HMMs. 
PRISM also includes similar utilities for calculating the
probability of a derivation or set of such and machine learning
algorithms which produce the most likely probabilities for switch
outcomes in order to explain a set of observed goals. 

\subsection{A framework for CHMMs in PRISM}
\label{sec:implementation}

We have implemented a framework for integration of side-constraints
in a PRISM program.\footnote{The current implementation of the framework is available via
http://akira.ruc.dk/$\sim$cth/chmm}
The framework has been used for adding
constraints to HMM based models, but it should be possible to extend
to other kinds of models. The underlying idea is that the program is
augmented with a constraint store and a constraint checker goal
is inserted in a few strategic places of the PRISM program.
This constraint checking is the direct implementation of check\_constraints 
of \emph{trans\_ctr}. The \emph{prune\_ctr} implementation is not discussed
as we use the tabling mechanism of PRISM to prune the search
space.

\subsubsection{Integration of side-constraints in a PRISM program}

This section describes how our framework can be integrated in
a PRISM program. As an example, we consider an implementation 
of the HMM from the previous section. Below the central 
recursive predicate of the implementation is shown extended
with constraint checking,

\begin{listing}[1]{1}
hmm(T,State,[Emit|EmitRest],StoreIn) :-
   T > 0,
   msw(trans(State),NextState),
   msw(emit(NextState),Emit),
   check_constraints([NextState,Emit],StoreIn,StoreOut),
   T1 is T-1,
   hmm(T1,NextState,EmitRest,StoreOut).
\end{listing}
Integration of side-constraint checking is done by extending relevant
predicates with an extra parameter (\texttt{StoreIn,StoreOut} in the
code above) to accommodate a constraint store and a call to the 
\texttt{check\_constraints} goal (line 5), after each distinct
sequence of \texttt{msw} applications. 

If \texttt{check\_constraints} fails during PRISM inference, then
the corresponding PRISM derivation fails, and further
extensions of this derivation will not be attempted since it does not
constitute a valid run. In effect, inference by PRISM will only consider runs which
are guaranteed not to violate any of the constraints declared for the model.

Declaration of constraints and implementation of constraint solvers are conceptually
decoupled from the PRISM model. The declaration of side-constraints on
the model is done by declaring facts of the form,
\texttt{constraint(ConstraintSpec)}.
The \texttt{ConstraintSpec} associates the constraint with a constraint checker
implementation and may contain some parameters for this particular
instance of the type of constraint.

A satisfiability checker maintains its own constraint store.
A satisfiability checker for a particular type of constraint consists
of an \texttt{init\_constraint\_store/2} rule and one or more
\texttt{check\_sat/4} rules. The
\texttt{init\_constraint\_store/2} rule is used to create a starting 
point for the constraint store of each declared constraint and is 
of the form,
\begin{verbatim}
init_constraint_store(ConstraintSpec, InitialStore).
\end{verbatim}
It is given \texttt{ConstraintSpec} and must unify
\texttt{InitialStore} with an initial constraint store matching the
\texttt{ConstraintSpec}. Additionally, one or more
\texttt{check\_sat} rules of the form,
\begin{verbatim}
check_sat(ConstraintSpec,StateUpdate,StoreBefore,StoreAfter):- ... .
\end{verbatim}
\noindent
must be implemented to check the satisfiability of the constraint.

As an example, consider an implementation of a \texttt{cardinality\_atmost}
constraint, called \verb+cardinality+ in our framework,
\begin{verbatim}
init_constraint_store(cardinality(_,_), 0).
check_sat(cardinality(U,Max), U, VisitsIn, VisitsOut) :-
        VisitsOut is VisitsIn + 1,VisitsOut =< Max.
check_sat(cardinality(X,_),U,S,S) :- X \= U.
\end{verbatim}

Each time
\texttt{check\_constraints} is called from the PRISM model, the relevant
\texttt{check\_sat} goals are called for each declared
constraint. If any of these fails, so will \texttt{check\_constraints}.
\texttt{StateUpdate} and \texttt{StoreBefore} are given and
\texttt{check\_cons- traints} is expected to unify \texttt{StoreAfter} to 
an updated constraint store. In our example HMM, the \texttt{StateUpdate}
will consist of the \texttt{[State,Emit]} pattern given to
\texttt{check\_constraints}.

The call to this rule must only succeed if the constraint given by
\texttt{ConstraintSpec} is not violated by the further information
given by the \texttt{StateUpdate}. Constraints are checked
incrementally and should only fail if any further updates to the
constraint store can only lead to failure. 

The constraint stores of individually declared constraints are
automatically aggregated in the constraint store exposed to
the PRISM model. Individual constraint checkers are unaware
of each other and cannot access the individual constraint stores
of other constraint checkers. The constraints are checked in the
order they are declared, so this order should be optimized to 
do pruning as early as possible.

\subsubsection{Efficient inference with a separate constraint store stack}\label{sec:separate_constraint_store}

The tabling mechanism in PRISM makes Viterbi computation and EM learning
efficient, but when extra parameters such as the constraint store are
introduced in the probabilistic goals, PRISM considers these as
goals with distinct derivations and stores a tabled entry for
each version of the goal. This behavior is undesired when the
extra parameters are used only for internal bookkeeping. The 
effect of this excessive tabling is that the dynamic programming
advantages are lost with exponential time inference as consequence.

In \cite{CG09} a related problem concerning tabling of annotations produced 
by running Viterbi on PRISM programs is approached using a program transformation that removes 
non-discriminating arguments, which do not affect the control flow. The annotation
can then be recovered from the program derivation of the transformed program. 

This approach is not applicable for the constraint store argument because the
constraint store implicitly affects control flow by limiting possible future derivation extensions. 
The constraint store has to be considered in the inference process; otherwise it would be possible
to produce invalid derivation paths. 

B-Prolog, on which PRISM is based, supports table modes, but this is not directly
usable  with probabilistic goals in PRISM. It is possible with these modes to declare
an argument of a tabled goal as an \emph{output argument}, which means that it will not be used as key in the table lookup, 
but will be unified with the value of the argument stored in a tabled goal.
For our purpose, declaring the constraint store arguments as output arguments would not be feasible
since different derivations of the same goal may have differing constraint stores and these
determine possible derivation extensions.

To deal with the tabling problem we have introduced a separate
constraint store stack, which avoids storing data locally in parameters of
probabilistic goals by maintaining the constraint store
with assert and retract. This stack is maintained in parallel to the
derivation stack of Prolog. PRISM utilizes Prolog's backtracking to explore possible
solutions, so the constraint store stack implementation is required to 
be able to restore a previous constraint store when PRISM encounters failures during inference
and performs backtracking to find alternative solutions. 

To utilize this functionality, the user should use the goal
\texttt{check\_constraints/1}, which omits the store arguments, rather
than \texttt{check\_constraints/3} as stated above. We then define \texttt{check\_constraints/1} as

\begin{verbatim}
check_constraints(StateUpdate) :-
    get_store(StoreBefore),
    check_constraints(StateUpdate,StoreBefore,StoreAfter),
    forward_store(StoreAfter).
\end{verbatim}

\noindent
The new \texttt{check\_constraints/1} make use of the goal \texttt{get\_store/1} to 
retrieve the current version of the constraint store and \texttt{forward\_store/1} is 
used to assert the updated store, 

\begin{verbatim}
get_store(S) :- !, store(S).
forward_store(S) :- (asserta(store(S)) ; retract(store(S)),fail).
\end{verbatim}

\noindent
If a derivation fails, PRISM backtracks to the choice point in 
the \texttt{forward\_store} rule and retract the most recently asserted
store. Then, when exploring alternative derivation extensions, the
previously asserted constraint store will be used as expected.

\subsubsection{Complexity analysis of  our implementation}

Due to tabling, PRISM guarantees familiar best known complexity bounds
of common inference tasks on a variety of the models that can be
expressed in PRISM, which includes HMMs \cite{Sat00}.
This implicitly limits the number of calls of
\texttt{check\_constraints} to the same bound.
The added complexity of doing constraint checking depends on  incremental
constraint checking cost of individual constraints checkers and the number of
constraints expressed on the model. 

Space complexity is influenced by table space usage and 
maximal length of a derivation at any given point. 
Since the asserted constraint store stack contains a constraint store fixpoint
for each step of the current derivation, it is bounded by
$O(n \text{max}(|c|))$ where $n$ is the length of the sequence and
$\text{max}(|c|)$ is the maximal size of the constraint store in any derivation step.
Note that the space complexity of the separate constraint
store stack is unaffected by time complexity and the number of states in the model.
With more complex models like the pair HMM, the table space
required for dynamic programming becomes the dominating concern.

\vspace{-0.2cm}

\section{Experimental validation}\label{sec:experiment_validation}

In this section, we validate our CHMM implementation with the pair HMM
presented in section \ref{sec:example}. The experiments were run on a computer 
with 16 2.4 GHz, 64 bit Intel Xeon(R) E7340 CPUs and 64 GB of memory. All of the experiments 
utilized only a single processor at a time. 

Our experiments utilize implementations of some common constraints
adapted for the CHMM framework: 
\texttt{cardinality(UpdatePatterns,Max)}
ensures that entries from the list \texttt{UpdatePatterns} occurs at most \texttt{Max} times
in the derivation sequence. 
\texttt{alldiff} ensures that all updates in a derivation are
different; \small \texttt{lock\_to\_sequence(Seq)} \normalsize  ensures that the sequence of
derivation updates is identical to the sequence represented by the
list \texttt{Seq}; \texttt{lock\_to\_set(Set)} ensures
that all updates belong to members of the list \texttt{Set}.
The operator \texttt{forall\_subseq(L,C)} applies the constraint C
to every subsequence of length L in the derivation sequence and
\texttt{for\_range(From,To,C)} applies C only the range,
\texttt{To}-\texttt{From}, both inclusive;
\texttt{state\_specific(C)} applies C only to the \texttt{State} part
of the update.

\subsection{Running time of constrained alignment}

The addition of side-constraints to an HMM involves  
some computational overhead in order to check the satisfiability of 
the constraints, but may also reduce the number of possible
solutions and therefore the amount of work required
to find the optimal path. As a practical experiment to 
demonstrate this, we consider global alignment with the 
pair HMM discussed in section \ref{sec:example}.

\begin{figure}[htb]
\vspace{-0.2cm}
\centerline{
 \includegraphics[scale=0.32]{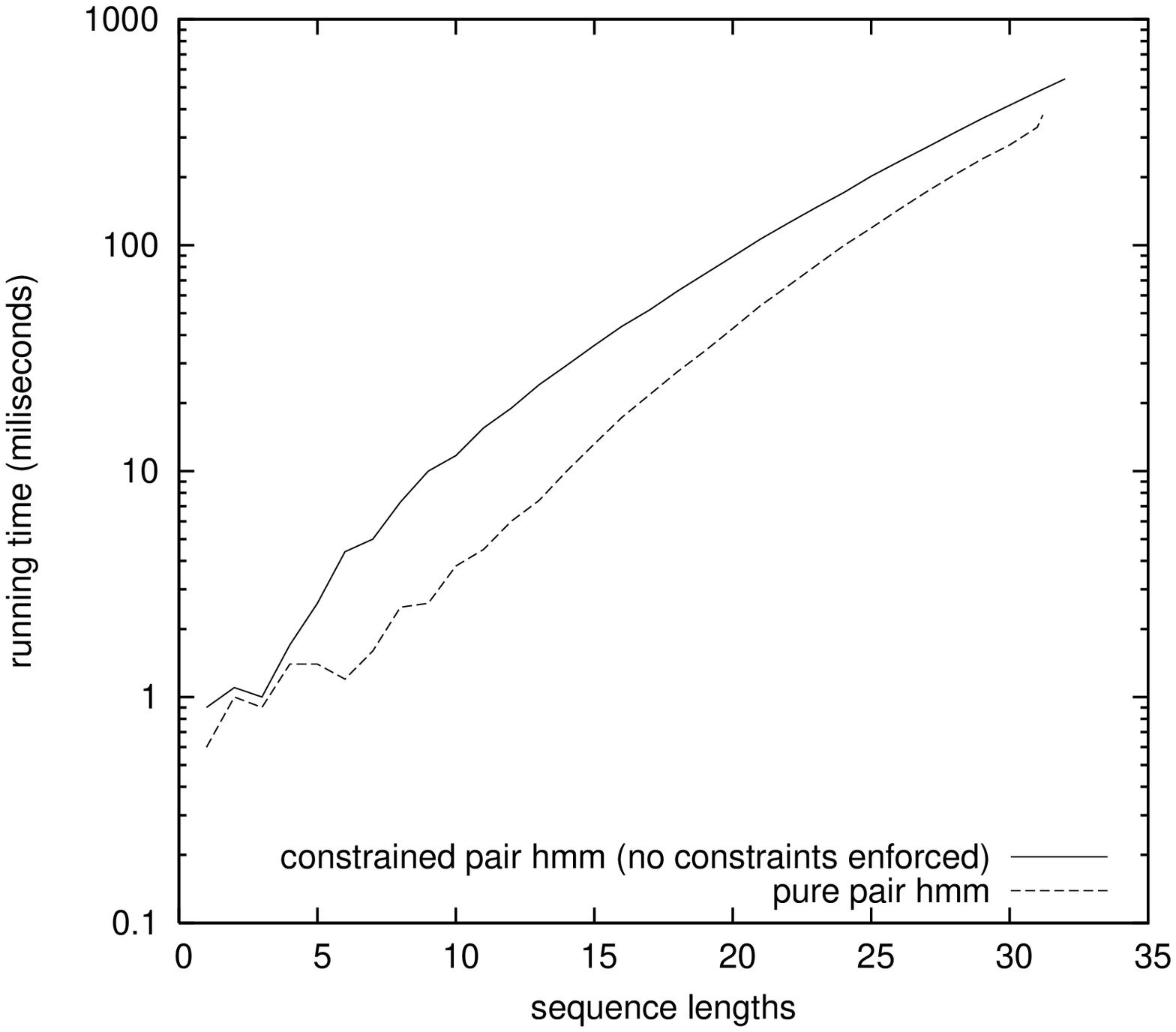}
 \includegraphics[scale=0.32]{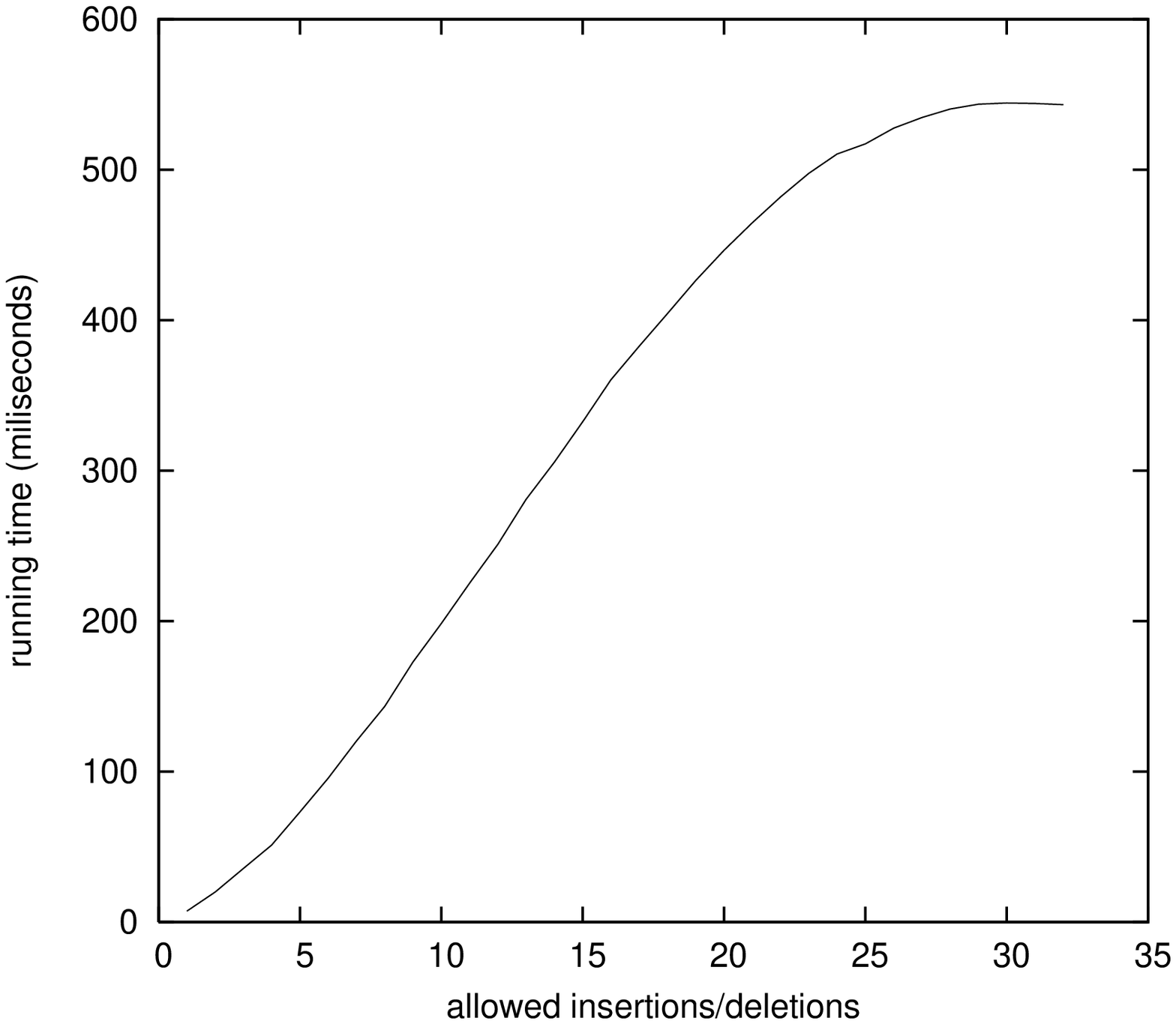}
}
\caption{
Left:  Running time of alignment with a pure pair HMM compared to alignment with a CHMM with no constraints enforced. Right: Running time of alignment of two sequences of length 32 with varying amounts of allowed insertions and deletions.}\label{fig:constrained_alignment_runtime}
\vspace{-0.2cm}
\end{figure}

The overhead of integrating the constraint checking 
machinery in the model is demonstrated in the left part of
Fig. \ref{fig:constrained_alignment_runtime}, where sequences of 
increasing length are aligned. It can be observed that the running time penalty
is a constant factor and that the polynomial time complexity of the pair HMM is preserved in our framework.
Obviously, polynomial time inference 
presupposes incremental constraint checking to be a constant time operation, 
which may not be the case for certain types of constraints.

In the right part of 
Fig. \ref{fig:constrained_alignment_runtime}, two sequences of equal
length (32) are aligned, but with varying amounts of constraints being enforced. The global cardinality
constraint is used to enforce an upper limit, \texttt{L}, on the
amount of inserts or deletes in the alignment,
\begin{verbatim}
constraint(state_specific(cardinality([insert,delete],L))).
\end{verbatim}

By constraining the alignment (allowing fewer gaps), the space of viable solutions is reduced.
The more constrained the alignment is, the more pruning opportunities arise. 
With a large amount of pruning opportunities,  the running time is reduced quite significantly.
Note that, since the imposed constraint is \texttt{state\_specific},
the number of possible alignments, and hence running time, is unaffected by
input sequence structure.

\subsection{Efficiency of the separate constraint store stack}

\begin{figure}[htb]
\vspace{-0.2cm}
\centerline{
 \includegraphics[scale=0.32]{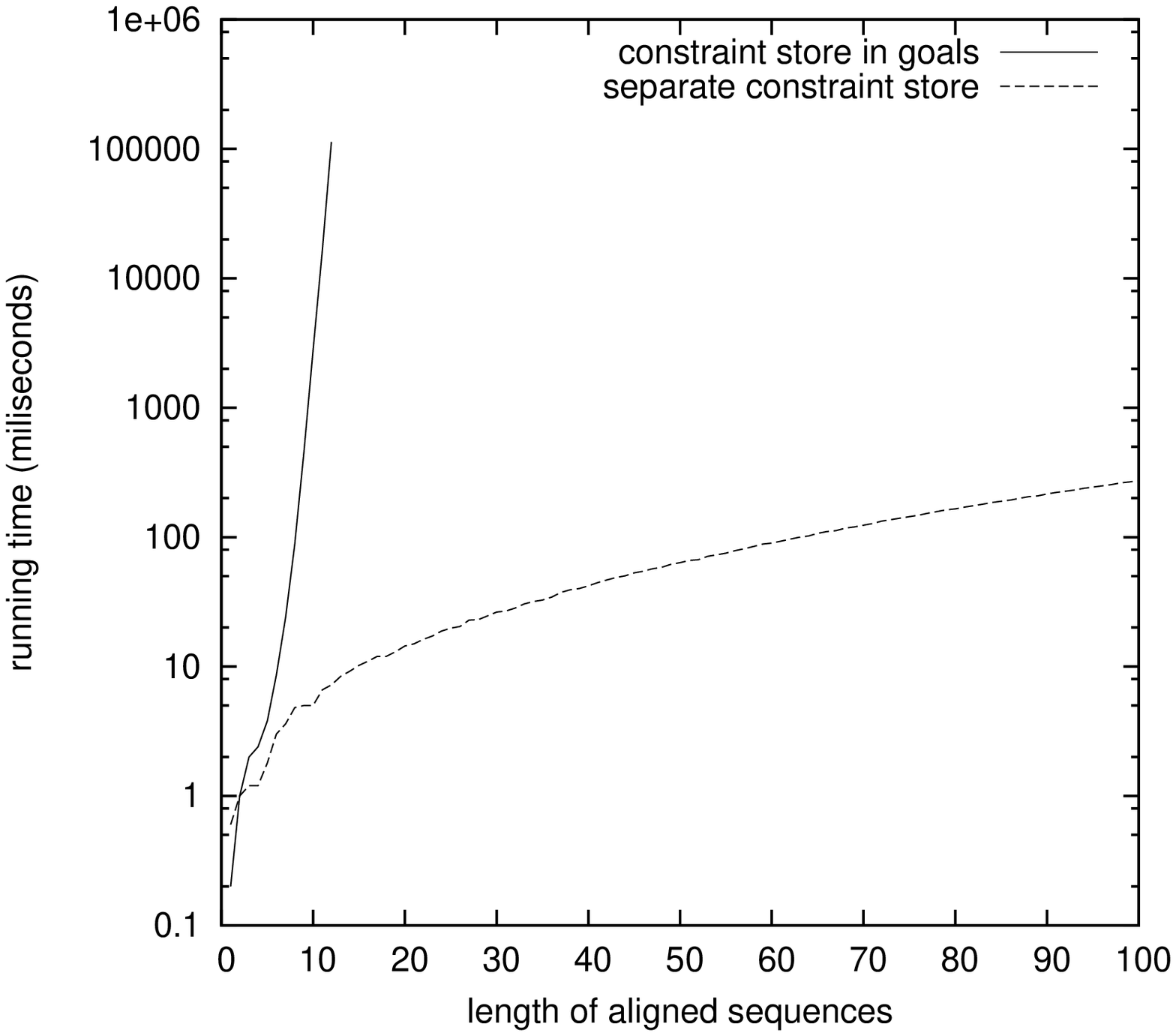}
 \includegraphics[scale=0.32]{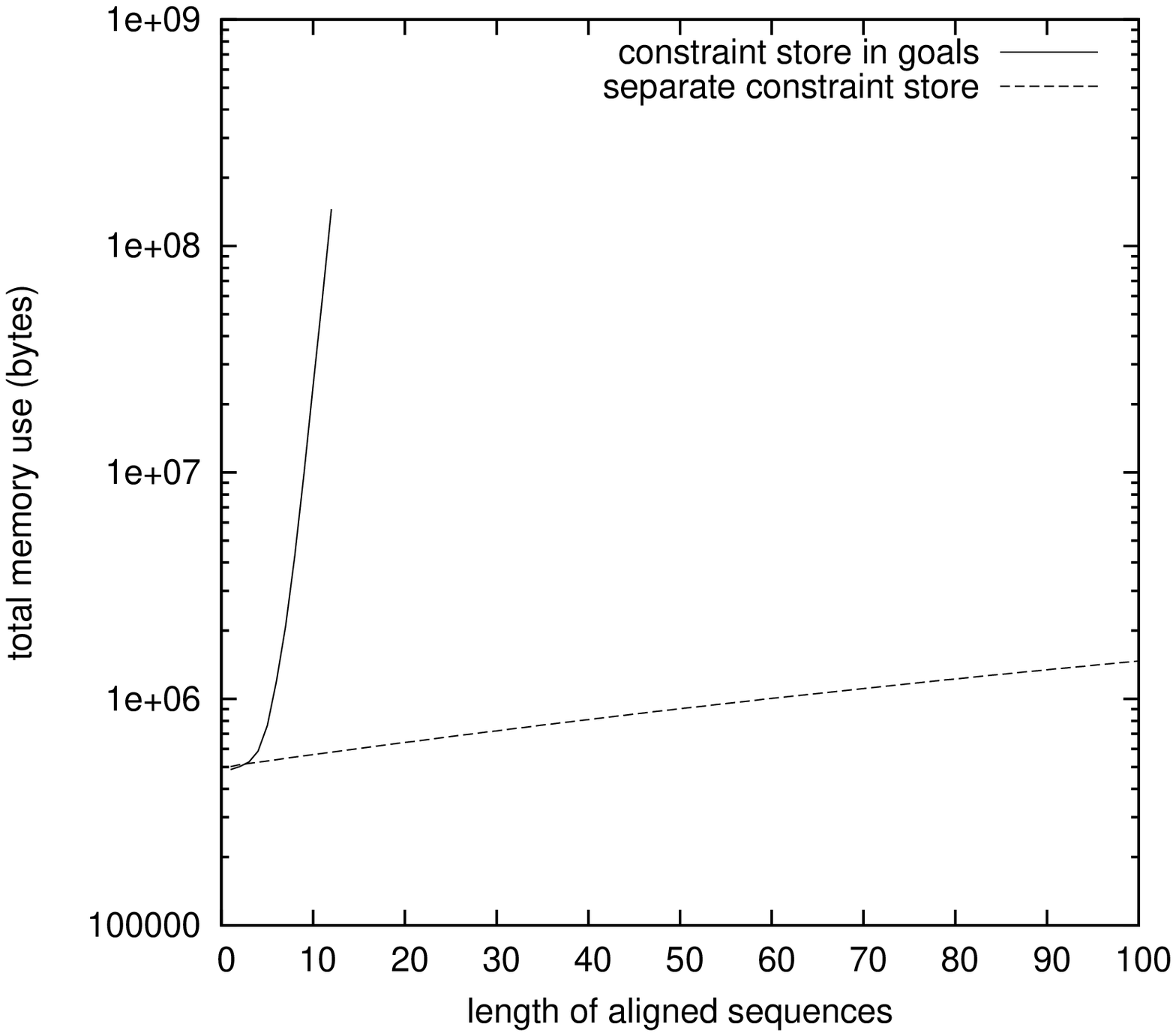}
}
\caption{A comparison of the running time (left) and memory usage
  (right) of constrained alignment of two sequences with tabled constraints
  versus a separate constraint store stack.}\label{fig:local_vs_global_store}
\vspace{-0.2cm}  
\end{figure}

To verify the efficiency of our constraint store implementation, 
alignment with a local cardinality constraint was measured for different sizes of input
sequences.
From the measurements, which are reported in
Fig. \ref{fig:local_vs_global_store}, it is apparent that our implementation does not incur the same exponential overhead
as the naive implementation where the constraint store is maintained in the goals and hence tabled.

Running times and memory usage for a range of different constraints 
are reported in Table \ref{table:constraint_running_times}. For the
sake for completeness, the table also includes running times for the
version where the constraint store is tabled.

\def\hack{\hbox to 0pt{\vbox to 12pt{\hbox to 0pt{\hss\vrule width  8.5pt height 0.5pt}\vfil}}}
\def\tspacer{\vrule width 0pt height 8pt}

\begin{table}[htb]
\footnotesize
\begin{tabular}{|p{4.5cm}|c|c|c|c|c|}
\cline{1-6}
 										 & Sequence&\multicolumn{2}{c|}{Running} &\multicolumn{2}{c|}{Memory} \\ 
 Constraint          & lengths &\multicolumn{2}{c|}{Time (in ms)} &\multicolumn{2}{c|}{consumption (in kb)}\\ \cline{3-6}
           &                   &\hack in goals  & separate &in goals  & separate           \\ \cline{1-6} 
\tspacer 
cardinality([insert],20) & 50 & 15460 & 3176 & 42296 & 5723 \\ \cline{1-6}
\tspacer
cardinality([insert],40) & 50 & 29557 & 3968 & 93845 & 6703          \\ \cline{1-6}
\tspacer
for\_range(1,50, lock\_to\_set([match])) & 100 & 24649 & 4544 & 105498 & 7137\\ \cline{1-6}
\tspacer
for\_range(1,90, lock\_to\_set([match])) & 100 & 20 & 48 & 1641 & 1198 \\ \cline{1-6}
\tspacer
for\_range(1,50, lock\_to\_sequence([match,..,match])) & 100 & 24829 & 4544 & 1641 & 1198 \\ \cline{1-6} 
\tspacer
for\_range(1,90, lock\_to\_sequence([match,..,match])) & 100 & 20 & 48 & 105498 & 7137 \\ \cline{1-6}
\tspacer
alldiff & 20 & 100442 & 28 & 85654 & 256 \\ \cline{1-6} 
\tspacer
forall\_subseqs(5,alldiff) & 10 & 1664 & 12 & 60098 & 137\\ \cline{1-6}
\end{tabular}
\caption{Running time and memory consumption for alignment with different kinds of constraints.}
\label{table:constraint_running_times}

\vspace{-0.2cm}
\end{table}

In most cases the separate constraint store performs better in terms of 
both running time and memory consumption. In the cases where
performance is worse, it can be attributed to a very small
number of possible derivations or constraints which rarely change the store.

\vspace{-0.2cm}
\section{Related work}\label{sec:related_work}

The term ``Constrained HMM'' is used in \cite{Row99,LME07} and
refers to restrictions on the finite automaton associated with an HMM
but not as constraints on HMM runs. In \cite{SK08}, CHMMs were introduced 
to exemplify an EM algorithm, suited for PRISM programs which allow the possibility of derivation failures.
Our approach differs, as we augment PRISM programs with side-constraints 
and use constraint solving techniques to achieve efficient inference.

In \cite{Rie98}, Riezler proposes techniques for inference in probabilistic constraint
logic programming. In \cite{CPC08} relationships between elements of a Bayesian Network are expressed as a constraint logic 
program, which is similar to the way we define HMMs.
However, our paper focus differs as we study the interest of checking satisfiability of side-constraints during inference.

In the natural language processing community, recent work on Constrained Conditional
Models feature an approach similar to ours. Indeed, Constrained Conditional Models is a
general framework that augments inference and learning of conditional models with
declarative constraints \cite{CRR08}. However, inference is expressed as an
Integer Linear Programming problem \cite{RY05}. 
In this context, more expressive
constraints, such as \texttt{cardinality} or \texttt{all\_different}, can not be 
added on an HMM run. Moreover, our PRISM-based implementation allows us to define the HMM structure
separately from the side-constraints and use advanced constraint solving techniques.

\section{Conclusions}\label{sec:discussion}

In this paper, we propose a framework to define HMMs with side-constraints as a Constraint Logic program
extended by probabilistic choices. Constraint Logic Programming have advantages in terms of more compact 
expression of CHMMs. Inference computations are adapted for CHMMs and conditions for an efficient
computation are described. An implementation based on PRISM is proposed and well-known
constraints and operators have been demonstrated  for defining CHMMs. Finally, we experimentally 
validate our approach with a constrained pair HMM used for biological sequence alignment. 

As current work, we study how sampling and EM-learning can be adapted
for our CHMM framework. Indeed, sampling turns out to be problematic in probabilistic models with a large probability of derivation failure. 
In \cite{SKZ05}, Sato et al. address the problem of EM-learning with PRISM
programs that can fail and their methods are also applicable for our framework.

As further work, we plan to incorporate more advanced constraint solving techniques such as those used
in Weighted CSP \cite{LS04} in the framework. 
This approach would allow us to combine soft constraints solving and inference and express this as an optimization problem.
We also plan to deal with the restriction that individual constraint checkers do not share information in our framework, so
that we can benefit from some of the optimization techniques used by other constraint solvers.
We are working on extending the library of constraints that can be defined as side-constraints. \\
\vspace{-0.2cm}
\subsubsection*{Acknowledgment} This work is supported by the project "Logic-statistic
modeling and analysis of biological sequence data" funded by the NABIIT program
under the Danish Strategic Research Council. We thank the anonymous reviewers for their
interesting comments.

\footnotesize

\end{document}